\documentclass[letterpaper, 10 pt, conference]{ieeeconf}
\usepackage[dvipsnames]{xcolor}
\usepackage{soul}
\usepackage{tikz}
\usepackage{multirow}
\usepackage{graphicx}
\usepackage{booktabs}
\usepackage{bm}
\usepackage{hyperref}
\usepackage{moresize}
\hypersetup{
    colorlinks=true,
    linkcolor=red,
    filecolor=magenta,      
    urlcolor=blue,
    pdftitle={darkgs},
    pdfpagemode=FullScreen,
    }
\usepackage{algorithm}
\usepackage{algpseudocode}
\usepackage{mathtools}
\usepackage{amsmath}
\usepackage{amssymb}
\usepackage{amsfonts}
\usepackage{subcaption}
\usepackage{caption}
\usepackage{cleveref}
\usepackage{perl_acronyms}
\let\labelindent\relax
\usepackage[inline]{enumitem}
\usepackage{siunitx}
\usepackage[
    style=ieee,
    doi=false,
    isbn=false,
    url=false,
    eprint=false,
    backend=bibtex,
    natbib=true
    ]{biblatex}
\IEEEoverridecommandlockouts                            
\title{\LARGE  \textbf{RecGS}: \textbf{Re}moving Water \textbf{C}austic with \textbf{Rec}urrent \textbf{G}aussian \textbf{S}platting}
\author{Tianyi Zhang$^{1}$, Weiming Zhi$^{1}$, Kaining Huang$^{1}$, \\ Joshua Mangelson$^{3}$, Corina Barbalata$^{2}$, Matthew Johnson-Roberson$^{1}$
\thanks{$^{1}$ Robotics Institute, School of Computer Science,
        Carnegie Mellon University,
        Pittsburgh, PA 15213, USA
        {\tt\small tianyiz4@andrew.cmu.edu}}%
\thanks{$^{2}$ Department of Mechanical \& Industrial Engineering,
        Louisiana State University,
        Baton Rouge, LA 70803, USA}
\thanks{$^{3}$  Department of Electrical and Computer Engineering, Brigham Young University,
Provo, UT 84602, USA}
}

\begin{document}
\maketitle
\thispagestyle{empty}
\pagestyle{empty}
\begin{abstract}
Water caustics are commonly observed in seafloor imaging data from shallow-water areas. Traditional methods that remove caustic patterns from images often rely on 2D filtering or pre-training on an annotated dataset, hindering the performance when generalizing to real-world seafloor data with 3D structures. In this paper, we present a novel method \emph{Recurrent Gaussian Splatting} (RecGS), which takes advantage of today's photorealistic 3D reconstruction technology, \ac{3DGS}, to separate caustics from seafloor imagery.
With a sequence of images taken by an underwater robot, we build \ac{3DGS} recurrently and decompose the caustic with low-pass filtering in each iteration. In the experiments, we analyze and compare with different methods, including joint optimization, 2D filtering, and deep learning approaches. The results show that our proposed RecGS paradigm can effectively separate the caustic from the seafloor, improving the visual appearance, and can be potentially applied on more problems with inconsistent illumination. 
Our videos and code are released at \url{https://tyz1030.github.io/proj/recgs.html}
\end{abstract}

\section{INTRODUCTION}
\label{sec:introduction}
With today's underwater robotic technologies, the benthic world has become more and more accessible to marine scientists, oceanographers, and the public. Robots can carry cameras and scan the seafloor in the form of images and videos. 3D computer vision techniques have been applied to recover the geometry and structure of the seafloor from these 2D imagery observations~\cite{gen2020mjr}. However, images taken underwater exhibit heavy water effects such as light attenuation, color distortion, and backscattering~\cite{jaffe}, which downgrades the visual appearance and the performance of the computer vision techniques being applied. \emph{Water caustics}, also known as sunlight flickering, is one of many water effects commonly observed. This refers to the phenomenon of shifting patterns on underwater surfaces, caused by light rays refracting through moving waves. This paper focuses on developing a 3D vision algorithm that removes the underwater caustic effect from a sequence of shallow-water seafloor observations. We seek to obtain photorealistic rendering of underwater scenes free of water caustics.
\par Within the literature of 3D reconstruction, \ac{SfM} methods are some of the most fundamental. These methods recover the camera poses and sparse 3D structure of the scene from a sequence of images~\cite{buildrome}. However, the visualization results of \ac{SfM} are generally sparse and not photorealistic. Therefore, recent advancements in 3D reconstruction such as \ac{NeRF}~\cite{mildenhall2020nerf} and \ac{3DGS}~\cite{kerbl3Dgaussians} build on top of the results of \ac{SfM} to enable photorealistic novel-view rendering of the scene. These methods constitute the \ac{SOTA} of 3D reconstruction. In general, \ac{SfM} pipelines are robust against illumination change due to sophisticated outlier rejection mechanisms. Therefore, they can be applied to internet data with different levels of ambient lighting and camera parameters~\cite{buildrome}, while still capable of accurately estimating the poses of the cameras. We also observe that water caustics generally have a limited impact on traditional \ac{SfM} pipelines for the same reason. Recovering photorealistic visualizations is of particular significance underwater and has implications for other branches of science. However, water caustics are detrimental to the construction of photorealistic representations, which are built on top of \ac{SfM}. These include
\ac{NeRF} and \ac{3DGS} which further learn the 3D scene by optimizing photometric loss instead of re-projection error. As a result, these models are critically affected by the illumination inconsistency from caustics. 
\begin{figure}[t]
  \centering 
  \begin{subfigure}[b]{0.98\linewidth} 
    \includegraphics[width=\textwidth]{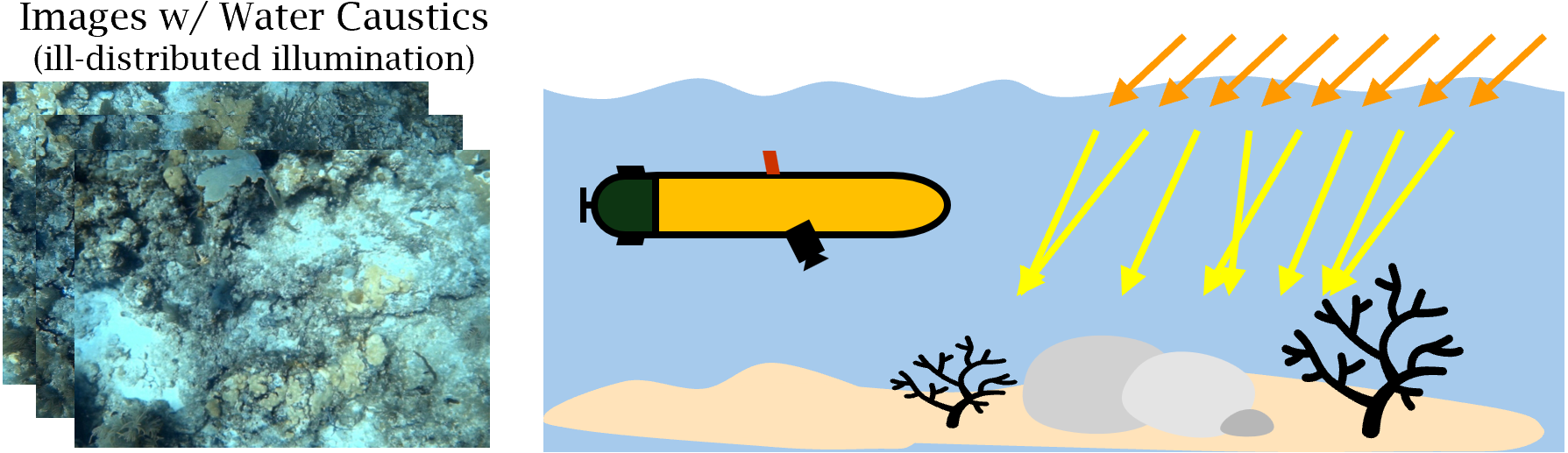}
    \caption{Underwater in shallow water is affected by caustics due to refraction from wave surface.}
    \label{intro:teasera}
  \end{subfigure}
  \vfill 
  \vspace{0.5cm}
  \begin{subfigure}[b]{0.98\linewidth} 
    \includegraphics[width=\textwidth]{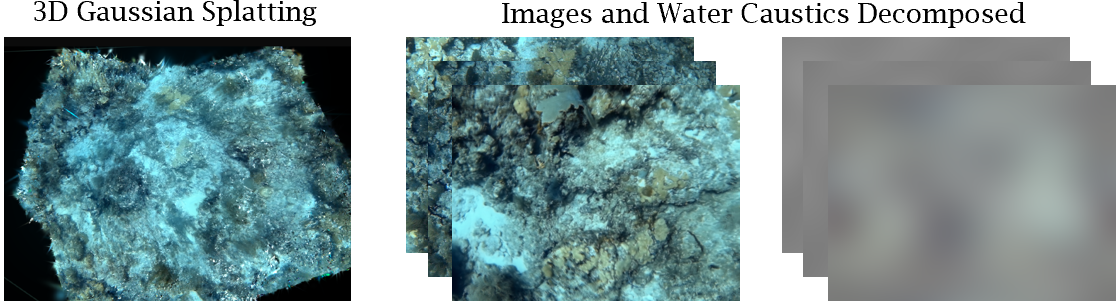}
    \caption{Our method recurrently builds 3DGS and approximates caustics with 2D Fourier transformation.}
    \label{intro:teaserb}
  \end{subfigure}
  \caption{Underwater imaging in shallow water suffers from water caustics. Our proposed pipeline based on \ac{3DGS} and Fourier low-rank reconstruction decomposes the background from water caustic without supervision.}
  \label{fig:teaser}
\end{figure}

\begin{figure*}[t]%
\centering
\includegraphics[width=0.95\linewidth]{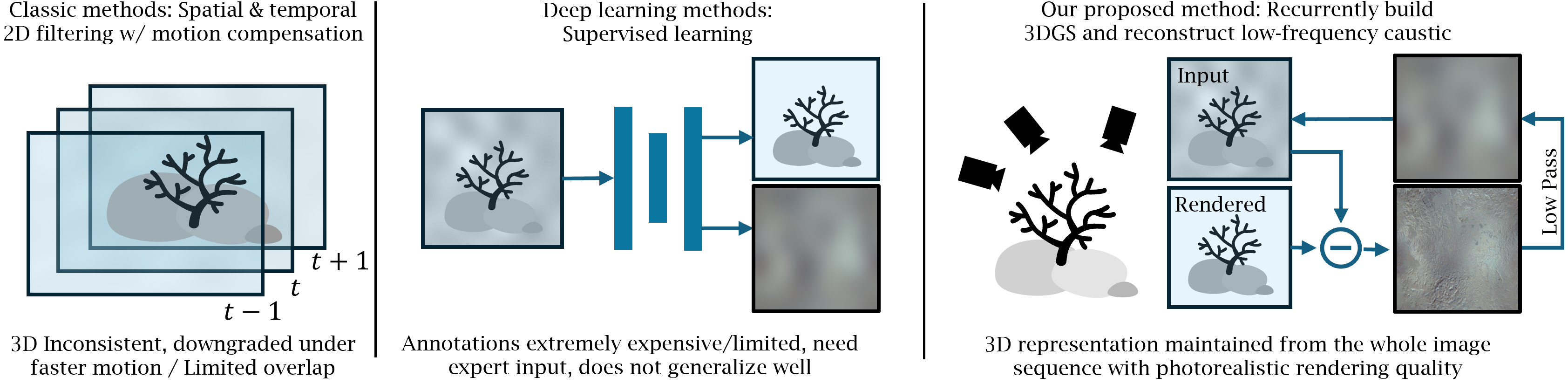}
    \caption{Related Works: \textbf{Classic methods} (Left) are based on image filtering along a certain time window on the 2D image space. The performance gets downgraded when a 3D structure is present in the scene, or if the camera moves fast and observes less overlap between frames. \textbf{Deep learning methods} (Mid) require expert annotations, which is extremely expensive to scale up. Neural networks trained with limited annotated data do not generalize well to novel observations. \textbf{Our proposed method} (Right) maintains dense 3D scene representations by building a 3D Gaussian model recurrently and decomposing low-frequency caustics from residuals. Our method works well on images captured from an underwater robot in the wild, without any pretraining on a dataset.}
    \label{relat:banner}
\end{figure*}
\par Existing studies on caustic removal are based either on image filtering or supervised deep learning. Filtering-based methods~\cite{ flicker04, filter2008oceans, flicker12} estimate and separate the caustic by running a non-linear temporal filter with motion compensation. However, such hand-crafted filtering techniques assume that the motion is small and has no awareness of 3D structure, thus the performance can be easily downgraded due to non-optimal hyperparameters, increased motion speed, and sealoor rugosity. Deep learning methods~\cite{deepcaustics2019, seafloorinv2023} require an annotated dataset on which to train. However, such dense caustic annotations are extremely expensive to acquire, hard to scale up, and only monochromatic~\cite{seafloorinv2023}. Therefore, methods based on deep learning perform poorly when generalizing to novel observations.
\par In this paper, we study caustic removal as a problem of illumination inconsistency within 3D structures. The community has seen related problems with color~\cite{neuralsea}, brightness~\cite{zhang2024darkgs} and exposure inconsistency~\cite{darmon2024robust, wang2024bilateral} being addressed with 3D vision techniques without any annotations. We seek to push the boundaries of this line of work and exploit the 3D consistency maintained in representation to remove caustics. To this end, we present \emph{Recurrent Gaussian Splatting} (RecGS), a novel framework which applies iterative low-pass filtering to refine the quality of the 3D representation during construction. Compared with existing approaches for water caustic removal, our proposed approach is unique in its structure-awareness and unsupervised nature. Namely, our approach leverages the consistency within the 3D structure of our representation to enhance performance and does not demand any annotated labels, enabling efficient scaling-up to large scenes. We provide extensive empirical evaluations on our underwater dataset, of shallow water corals, collected with a \ac{ROV} in Florida Keys, USA. Our method consistently and robustly produces clean photorealistic representations of underwater scenes that are caustic-free. A high-level illustration of our problem setup and method is provided in Fig.~\ref{fig:teaser}. 


\section{RELATED WORK}
\label{sec:relatedwork}
In this section, we discuss classic filtering-based methods in~\ref{subsec:classic} and Fig.~\ref{relat:banner} (left), deep learning-based methods in~\ref{subsec:dl} and Fig.~\ref{relat:banner} (mid), and \ac{NeRF} and \ac{3DGS} which are closely related to our work in~\ref{subsec:nerf},~\ref{subsec:rest} and Fig.~\ref{relat:banner} (right).

\begin{figure*}[t]%
\centering
\includegraphics[width=0.98\linewidth]{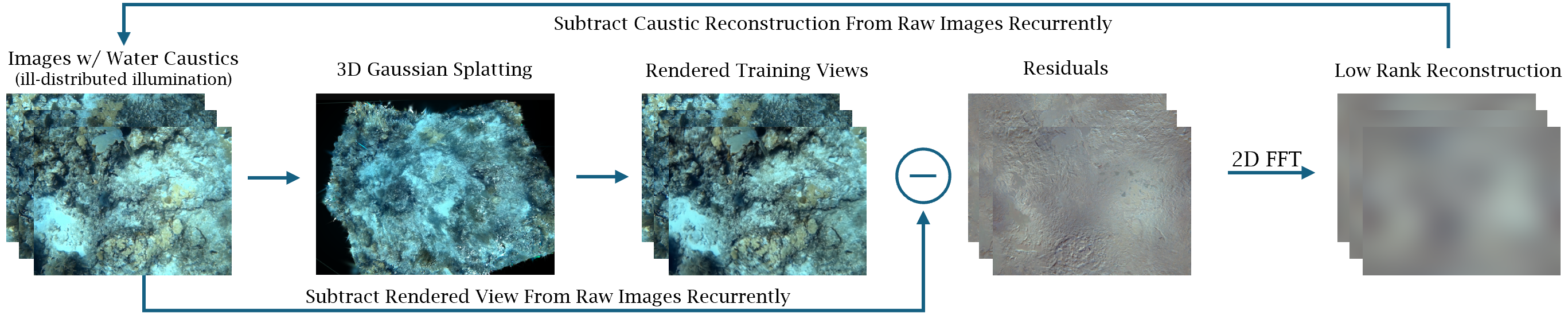}
    \caption{Our proposed recurrent \ac{3DGS} workflow: we build a vanilla \ac{3DGS} model with caustics in the images first, then find the residual between captured image and rendered view. We run 2D \ac{FFT} on the residual image, and reconstruct it with only the low-rank part. This low-rank reconstruction is then subtracted from the training images.}
    \label{method:banner}
\end{figure*}

\subsection{Classic caustic removal methods}
\label{subsec:classic}
Early work removes underwater caustic with a sequence of consecutive frames taken in a short period~\cite{flicker04}. Spatial derivatives are extracted from each image frame and temporal median filtering is performed to eliminate wide-band caustics. However, the camera motion is not considered in this approach. 
This method based on non-linear filtering is further developed in~\cite{filter2008oceans}, which leverages a similar setup that averages out the inconsistent illumination in a sequence of images, but with camera motion compensated.
Follow-up work~\cite{flicker12} enables online caustic filtering based on a similar principle, but based on the assumption that the camera motion is smooth and the seafloor is flat, hindering the generalizability on real-world images. An adaptive filtering strategy was proposed in~\cite{selfflicker15} to optimize the filter parameters in real time, but as acknowledged by the authors, the performance is still sensitive to hyperparameter tuning.
\par Overall, classic computer vision methods such as pose estimation and nonlinear filtering has been widely employed to compensate caustics in the water, but the operations are all conducted on 2D image space, which has limited capability for handling 3D structured observations and dealing with a fast moving camera with limited overlap from frame to frame.

\subsection{Deep learning-based caustic removal}
\label{subsec:dl}

Deep neural networks have shown great ability to learn the pixel-wise distribution in the 2D image space~\cite{unet}, with a potential to extend to different modalities~\cite{crosca} and physical phenomena~\cite{cai2021physics}. DeepCaustic~\cite{deepcaustics2019} proposed a pipeline formed of two CNN-based networks. The first network learns a saliency map, and the second learns to recover the image free of caustics. Stereo and multi-view vision is studied in ~\cite{seafloorinv2023} to remove caustics with a neural net while keeping the seafloor structure invariant in the restored image. However, all deep learning based methods mentioned above need to train the neural network on a dataset with ground truth annotations and masks, making it hard to be scaled up and perform poorly when being generalized to novel observations.

\subsection{NeRF and 3D Gaussian Splatting}
\label{subsec:nerf}
Constructing representations of their surroundings is central to both robots deployed outdoors ~\cite{occgm,dir_maps}, and those operating indoors~\cite{wright2024v, C_mapping}. The widely used \ac{SfM}~\cite{Ullman1979TheIO, buildrome} has facilitated photorealistic 3D reconstruction in recent years. There has been extensive work around both feature-based \ac{SfM}~\cite{schoenberger2016sfm, sarlin20superglue} and learning-based alternatives~\cite{wei2020deepsfm, dust3r_cvpr24}. Based on the camera pose estimation from \ac{SfM}, \ac{NeRF}~\cite{mildenhall2020nerf} models a 3D scene using a continuous function, e.g. a \ac{MLP}, and can synthesize novel-view images with photorealistic quality.
Later on, \ac{NeRF} has been improved in terms of rendering quality and efficiency by its descendants such as Mip-NeRF~\cite{barron2022mipnerf360} and Instant-NGP~\cite{mueller2022instant}.
To alleviate the heavy sampling load in rendering, \ac{3DGS}~\cite{kerbl3Dgaussians} replaced the neural network backbone with 3D Gaussians, allowing for a rendering speed of 100+ fps while the whole system remains differentiable.
In this work, we build 3D seafloor models based on \ac{3DGS}, and can thus reconstruct the images from designated views. Accurate environment representations are critical for various downstream robot decision-making~\cite{Diff_temp,rana2017towards, Skill_acquisition_ode} and planning tasks~\cite{LaValle1998RapidlyexploringRT, para_planning,Planning_geo}.

\subsection{Image restoration based on NeRF and 3DGS}
\label{subsec:rest}
Optimizing a 3D model and an image perturbation model together has been widely used. The work in~\cite{neuralsea} optimizes the color distortation and backscattering model together with a neural reflectance field model. In DarkGS~\cite{zhang2024darkgs}, the parameter of the artificial light source can be optimized with a \ac{3DGS} model. RobustGS~\cite{darmon2024robust} optimizes an affine transformation mode to correct the color inconsistency in the GS training data. An attempt to handle reflective surfaces is presented in~\cite{jiang2024gaussianshader}. Neural representations extend beyond statically representing structures. The robustness of neural representation based \ac{SLAM} under such visual perturbation has been systematically studied in~\cite{xu2024customizable}. The multi-view NeRF rendering techniques have also been applied to robot motion generation in~\cite{Dia_teaching, scene_rep}. In this work, we identify and solve the ill-posing problem in the joint optimization method (\ref{subsec:joint}) and propose a recurrent framework that robustly restores the image from underwater caustics.

\section{METHODOLOGY}
\label{sec:methodology}
\subsection{Preliminary}\label{meth:pre}

Given a consecutive image sequence (of seafloor observations), camera poses and a 3D keypoint cloud can be estimated from \ac{SfM} pipeline such as COLMAP~\cite{buildrome}. With \ac{3DGS}, a 3D Gaussian cloud $G$ is initialized with this 3D keypoint cloud.
\par For each camera pose in the image sequence, an image can be synthesized from the following rendering equation~\cite{kerbl3Dgaussians}:
\begin{equation}
\label{eq:rendereq}
    \hat{I} = \sum_{\substack{i\in\mathcal{N}}}c_i\alpha_i \overset{i-1}{\prod_{\substack{j=1}}} (1-\alpha_j)
\end{equation}
where $c_i$ and $\alpha_i$ are the RGB radiance and volume density of $i^{th}$ Gaussian in the 3D Gaussian model $G$. Given a training image $I$, $G$ is optimized as follows:
\begin{equation}
\label{eq:vanigs}
    \underset{G}{\text{minimize}}\  \mathbf{dist}(I, \hat{I}) 
\end{equation}
here $\mathbf{dist()}$ is the distance metric same with~\cite{kerbl3Dgaussians}. Optimizing $G$ refers to optimizing all the parameters including $c_i$ and $\alpha_i$ of each Gaussian.
\par Water caustic can be observed in shallow water as the natural light entering the water is refracted at different angles due to the fluctuating wave surface. The refracted light creates focused and defocused pattern on the sea bottom. While it is hard to model the exact physics, we approximate the caustic $C$ as an additive radiance from the scene (similar to previous studies~\cite{filter2008oceans}). So the optimization problem becomes:
\begin{equation}
\label{eq:reoptim}
    \underset{G}{\text{minimize}}\  \mathbf{dist}(I-C, \hat{I}) 
\end{equation}
Note here $C$ has the same image format as the training images, which has all R, G, and B channels. This is different from previous studies which mostly model the caustic as monochromatic~\cite{deepcaustics2019, seafloorinv2023}.

\subsection{Residual Reconstruction with 2D \ac{FFT}}\label{meth:residual}
From our observation, building a scene with vanilla \ac{3DGS} under water caustic will suffer from photometric inconsistency. The caustic pattern, which causes such an inconsistency, is reflected in the residual of the rendered image. We propose to approximate the caustic caused by the refraction from the wave surface with 2D low-rank Fourier series.
By building a vanilla \ac{3DGS} from the image sequence, the residual can be calculated by $R = I-\hat{I}$. We implemented a low-pass filter with 2D \ac{FFT} to approximate the low-frequency illumination inconsistency $C$ caused by caustics (this step is similar to the Butterworth filter used in~\cite{filter2008oceans}):
\begin{equation}
\label{eq:caustic}
    C = \texttt{ifft}(\texttt{fft}(R)_{[0:k]})
\end{equation}
Here $\texttt{fft}()$ and $\texttt{ifft}()$ are  2D discrete Fourier transform and its inverse operation. $[0:k]$ refers to that only the $k$ frequencies with the lowest absolute value in the frequency space are kept. We use $k=9$ which is chosen empirically.

\subsection{Recurrent \ac{3DGS}}\label{meth:recur}
\begin{algorithm}[t]
\caption{recurrent \ac{3DGS}}
\label{algo:recgs}
\begin{algorithmic}[1]
    \State \textbf{Input:} Images $I$, Camera poses
    \State \textbf{Output:} Rendered View $\hat{I}$, Caustic per view $C_{ret}$
    \Procedure{TrainrecurrentGS}{$I$}
        \State     $\underset{G}{\text{minimize}}\  \mathbf{dist}(I, \hat{I})$ with Eq.~\ref{eq:rendereq}
        \State $C_{ret} = \texttt{zeroslike}(I)$
        \While{True}
            \State $R = I-\hat{I}$
            \State $C = \texttt{ifft}(\texttt{fft}(R)_{[0:k]})$\ (Eq.~\ref{eq:caustic})
            \If{$\|C-C_{ret}\|_2<threshold$}
                \State  $C_{ret} = C$
                \State \textbf{break}
            \Else
                \State $\underset{G}{\text{minimize}}\  \mathbf{dist}(I-C, \hat{I})$ (Eq.~\ref{eq:reoptim})
            \EndIf
        \EndWhile
        \State Return $\hat{I}$, $C_{ret}$
    \EndProcedure
\end{algorithmic}
\end{algorithm}

While the caustic can be decomposed from the frequency spectrum according to the method in~\ref{meth:residual}, it does not completely eliminate caustic in the image sequence. In other words, the illumination inconsistency is only partially reflected in the residual. We propose a recurrent \ac{3DGS} framework (Fig.~\ref{method:banner}) that progressively removes caustics.
\par With the optimization result from Eq.~\ref{eq:vanigs}, an initial estimate of caustic $C$ can be solved from Eq.~\ref{eq:caustic}. Then the \ac{3DGS} model can be re-optimized with Eq.~\ref{eq:reoptim}. With an updated \ac{3DGS} model, Eq.~\ref{eq:caustic} can be updated. Here we find that Eq.~\ref{eq:reoptim} and Eq.~\ref{eq:caustic} can be solved alternatively until the estimated caustic $C$ converges. See Algorithm~\ref{algo:recgs} for details of recurrent training \ac{3DGS}. The algorithm will return the image $\hat{I}$ caustic-free, and caustic $C_{ret}$ which is reconstructed with low rank 2D \ac{FFT}.

\section{EXPERIMENTS}
\label{sec:experiments}
In this section, we provide rigorous empirical evaluations of our proposed RecGS method. We provide details of our experiment setup in~\cref{subsec:Exp_setup}. Then, seek to investigate and provide answers to the following questions:
\begin{itemize}
    \item Should the caustic effects be learned and optimized jointly with \ac{3DGS}? (\Cref{subsec:joint})
    \item Are pretrained deep-learning approaches sufficient to remove water caustics? (\Cref{subset:deep_learnining_fails})
    \item How does recurrent Gaussian Splatting compare with alternative filtering-based approaches? (\Cref{subsec:filter})
    \item How effective are the recurrences? (\Cref{subset:recurrences})
\end{itemize}

\subsection{Experiment Setup}\label{subsec:Exp_setup}
\subsubsection{Data Collection} 
\begin{figure}[t]
  \centering 
  \begin{subfigure}[b]{0.45\linewidth} 
    \includegraphics[width=\textwidth]{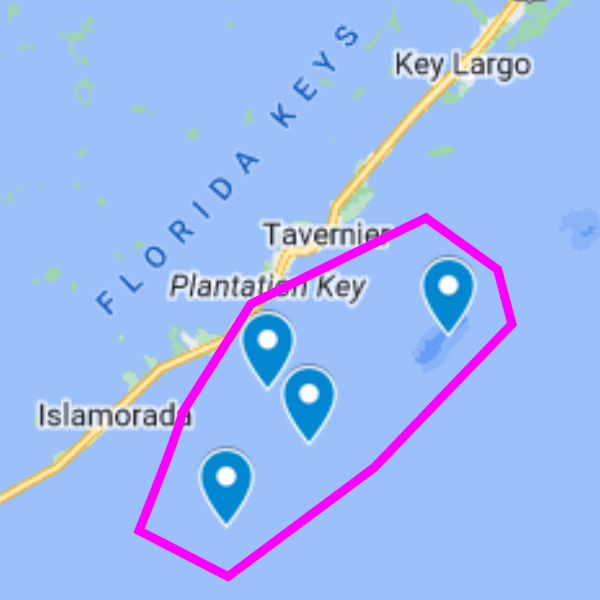}
    \caption{Data collection site.}
    \label{fig:deploya}
  \end{subfigure}
  \hspace{0.02\linewidth}
  \begin{subfigure}[b]{0.45\linewidth} 
    \includegraphics[width=\textwidth]{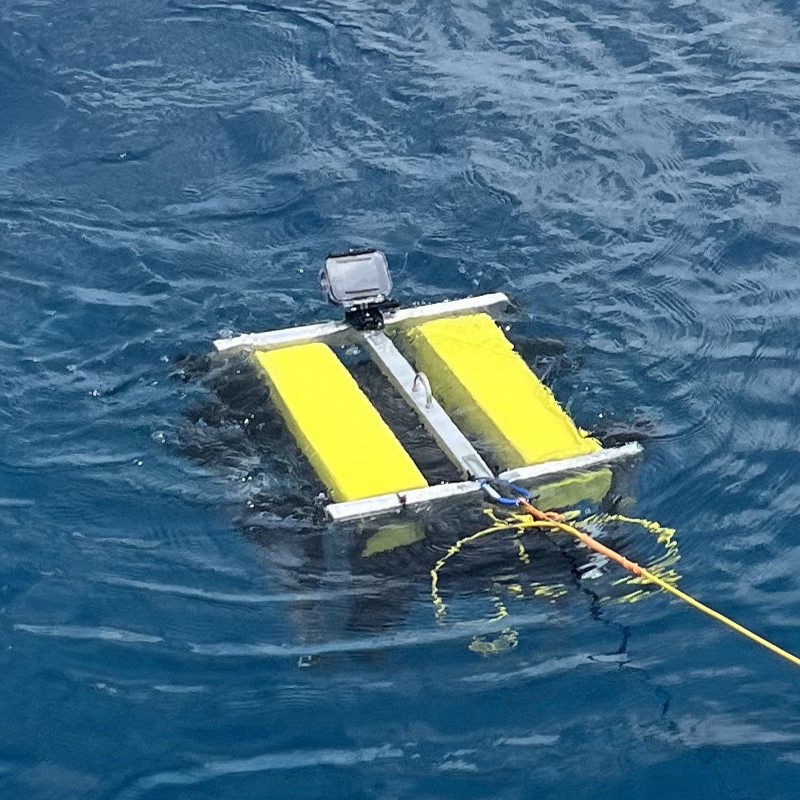}
    \caption{Robot platform.}
    \label{fig:deployb}
  \end{subfigure}
  \caption{We collected data in real world marine environment in Florida Keys area. The Robot we use is a LSU's Bruce \ac{ROV}~\cite{bruce2023} equipped with ZED cameras.}
  \label{fig:deploy}
\end{figure}

We collect a dataset of underwater images by deploying LSU's Bruce \ac{ROV}~\cite{bruce2023} equipped with a ZED camera in a coral reef sanctuary in Florida. Illustrations of the location and robot platform are provided in~\cref{fig:deploy}. The weather during data collection is sunny and the water depth during data collection ranges from 5 to 10 meters.
\subsubsection{Training} We run experiments on 4 selected image sequences with no obvious moving objects and complete COLMAP reconstruction with all the input images. We use the renderer implementation of \ac{3DGS}~\cite{kerbl3Dgaussians} as backbone and PyTorch \texttt{fft} module for filtering and caustic reconstruction.

\subsection{Should we jointly optimize caustic $C$?}
\label{subsec:joint}
The very first question we answer is that instead of decomposing the caustic from residual recurrently, should we jointly optimize per-frame caustic $C$ together with \ac{3DGS}? Consider 2D frequency response $\mathbf{c}_{[0:k]}$ which has lowest $k$ frequencies as variables to be optimized and the high-frequency part are all fixed to $0$. Then Eq.~\ref{eq:caustic} becomes the following:
\begin{equation}
    C = \texttt{ifft}(\mathbf{c}_{[0:k]})
\end{equation}
And then the optimization problem in Eq.~\ref{eq:reoptim} will optimize $G$ together with $\mathbf{c}_{[0:k]}$:
\begin{equation}
    \underset{G, \mathbf{c}_{[0:k]}}{\text{minimize}}\  \mathbf{dist}(I-C, \hat{I}) 
\end{equation}

\begin{figure}[t]%
\centering
    \includegraphics[width=0.95\linewidth]{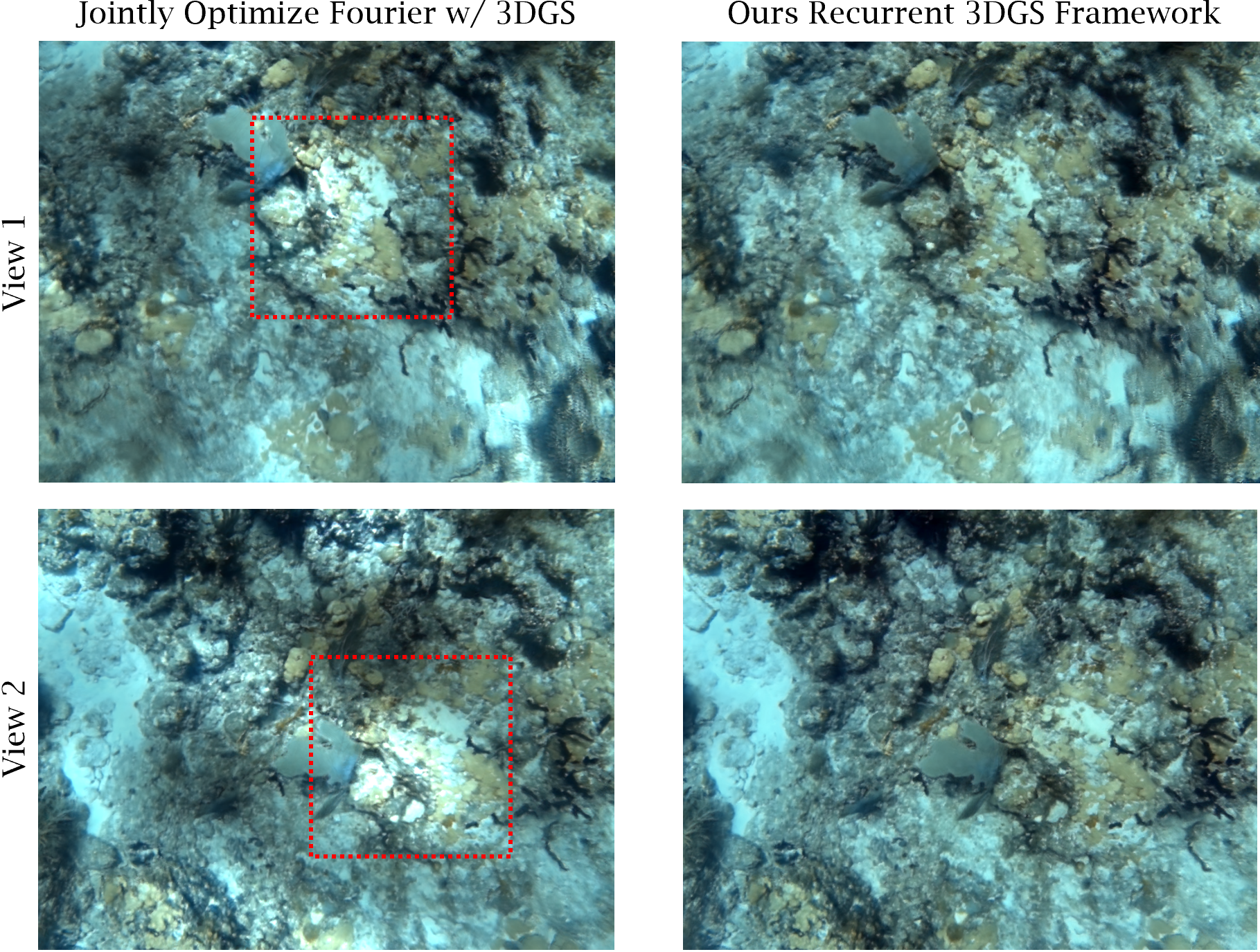}
    \caption{Jointly optimizing a low-rank Fourier spectrum together with \ac{3DGS} leads to an ill-posed behavior. As shown in \textcolor{red}{dashed box}, joint-optimization method creates undesired over-exposed areas, while still maintaining multi-view consistency. In comparison, our recurrent method restores the scene with uniform illumination.}
    \label{exp:diropt}
\end{figure}

The results are shown in Fig.~\ref{exp:diropt}. We can see that jointly optimizing caustic together with \ac{3DGS} will lead to ill-balanced illumination. In comparison, our proposed recurrent \ac{3DGS} framework behaves constantly to maintain a balanced illumination. We also find that when evaluating the results of the same scene from different views, both methods can maintain 3D consistency. This implies that while we can learn the caustic model with resort to 3D consistency, such consistency does not guarantee the good performance of the result. As shown in Fig.~\ref{exp:diropt}, the result of joint-optimization method is ``consistently wrong". We need to carefully engineer the pipeline to constrain the model converging towards the desired direction. In this study, our recurrent pipeline shows a great performance gain over the counterpart.

\subsection{Failure case of Pre-trained Deep Learning Methods}\label{subset:deep_learnining_fails}

\begin{figure}[t]%
\centering
    \includegraphics[width=0.95\linewidth]{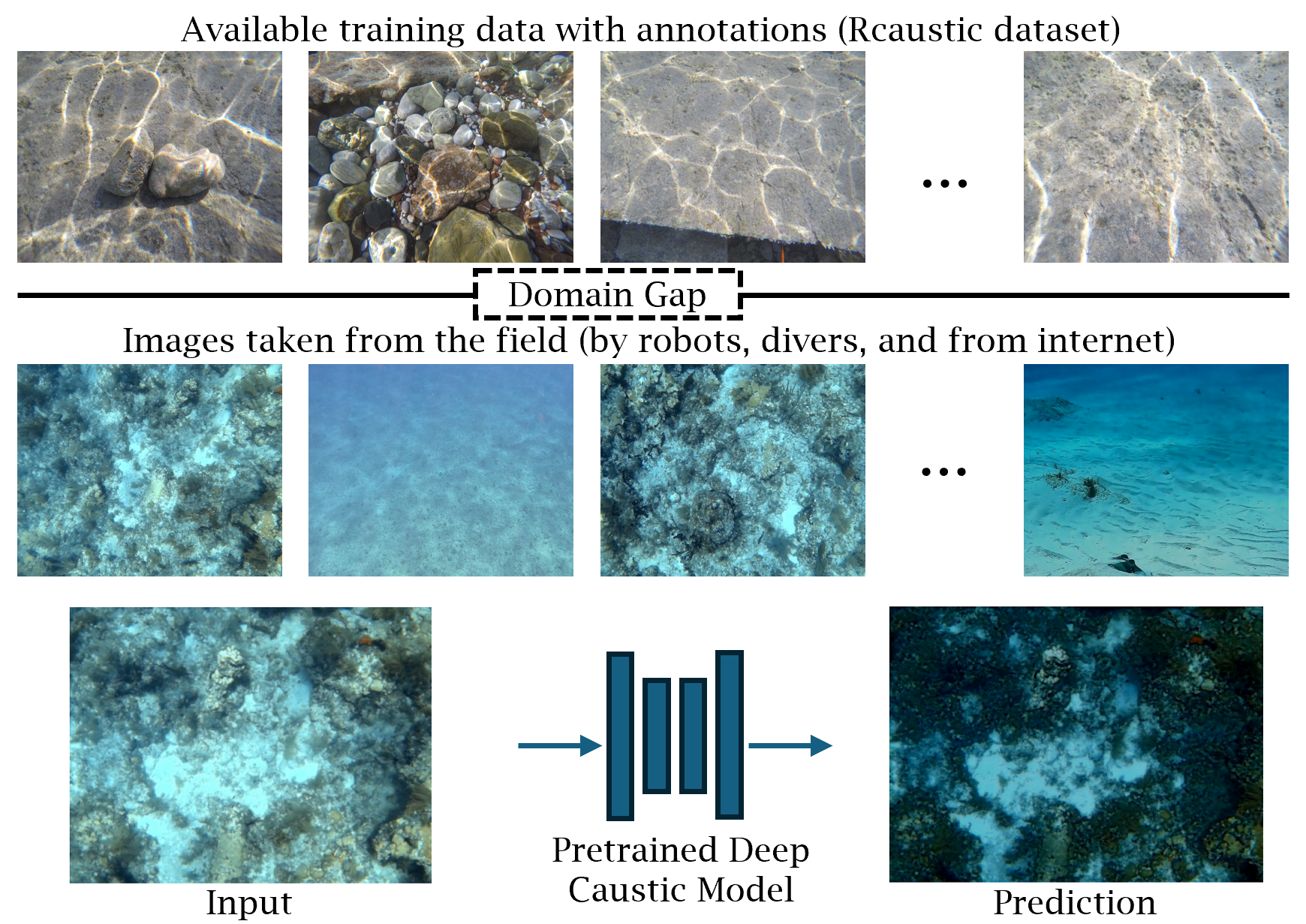}
    \caption{Failure of deep learning methods: There is a significant domain gap between the caustic removal training data nowadays and real world data collected from field deployment. Pre-trained deep neural networks can thus perform poorly when such domain gap presents.}
    \label{exp:deep}
\end{figure}

\begin{figure*}[t]%
\centering
\includegraphics[width=0.95\linewidth]{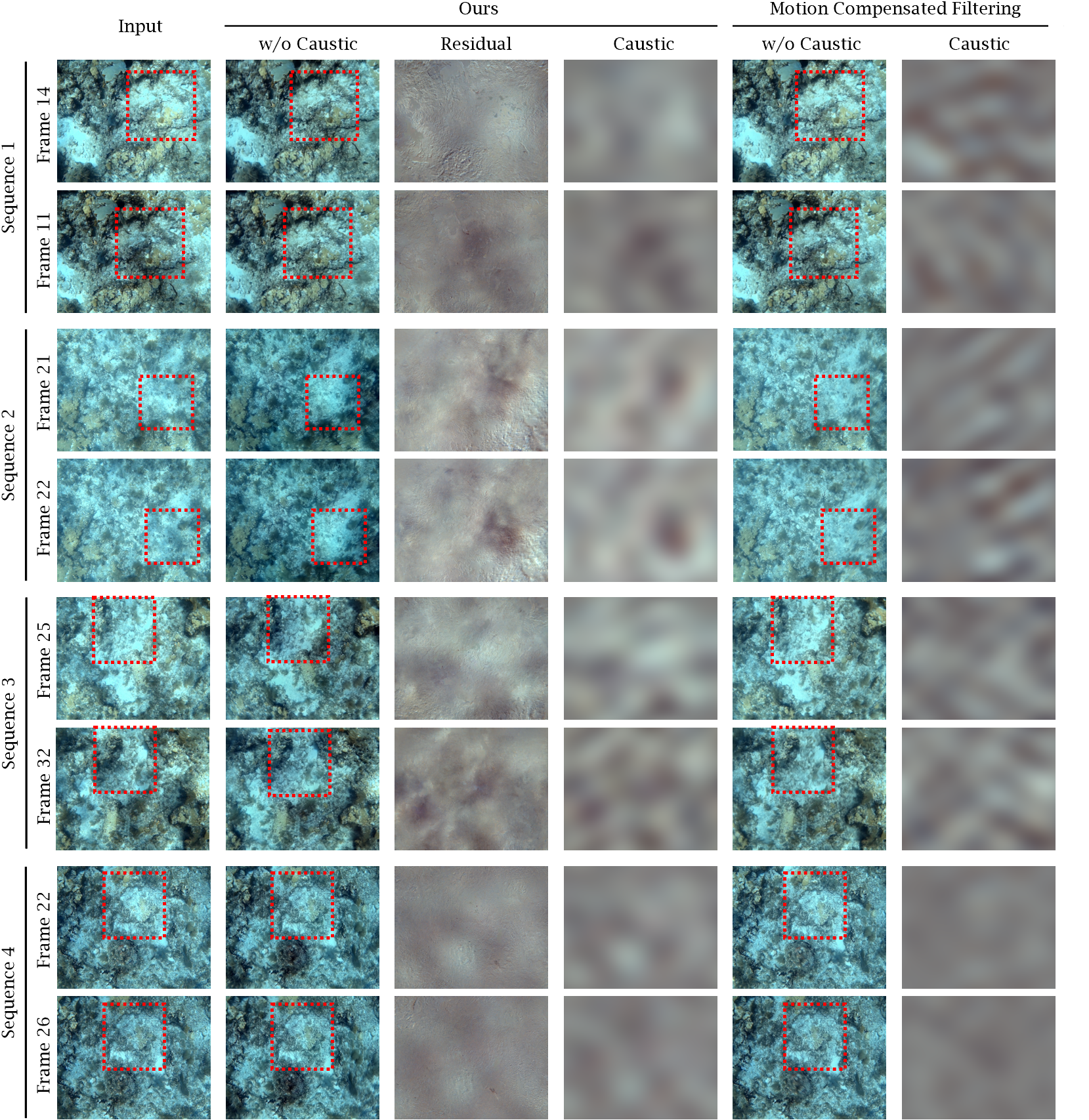}
    \caption{Visualization of results from multiple data sequences compared with motion-compensated filtering~\cite{filter2008oceans}: For each sequence, we picked up two frames with observable camera motion and caustics between them. We highlight the area with the most significant caustic variation from one frame to another in \textcolor{red}{dashed box}. From our results in column 2, we can see that the same areas in different frames are corrected with consistent illumination.
    }
    \label{exp:viz}
\end{figure*}

\begin{figure}[t]
  \centering 
  \begin{subfigure}{0.88\linewidth} 
    \includegraphics[width=\textwidth]{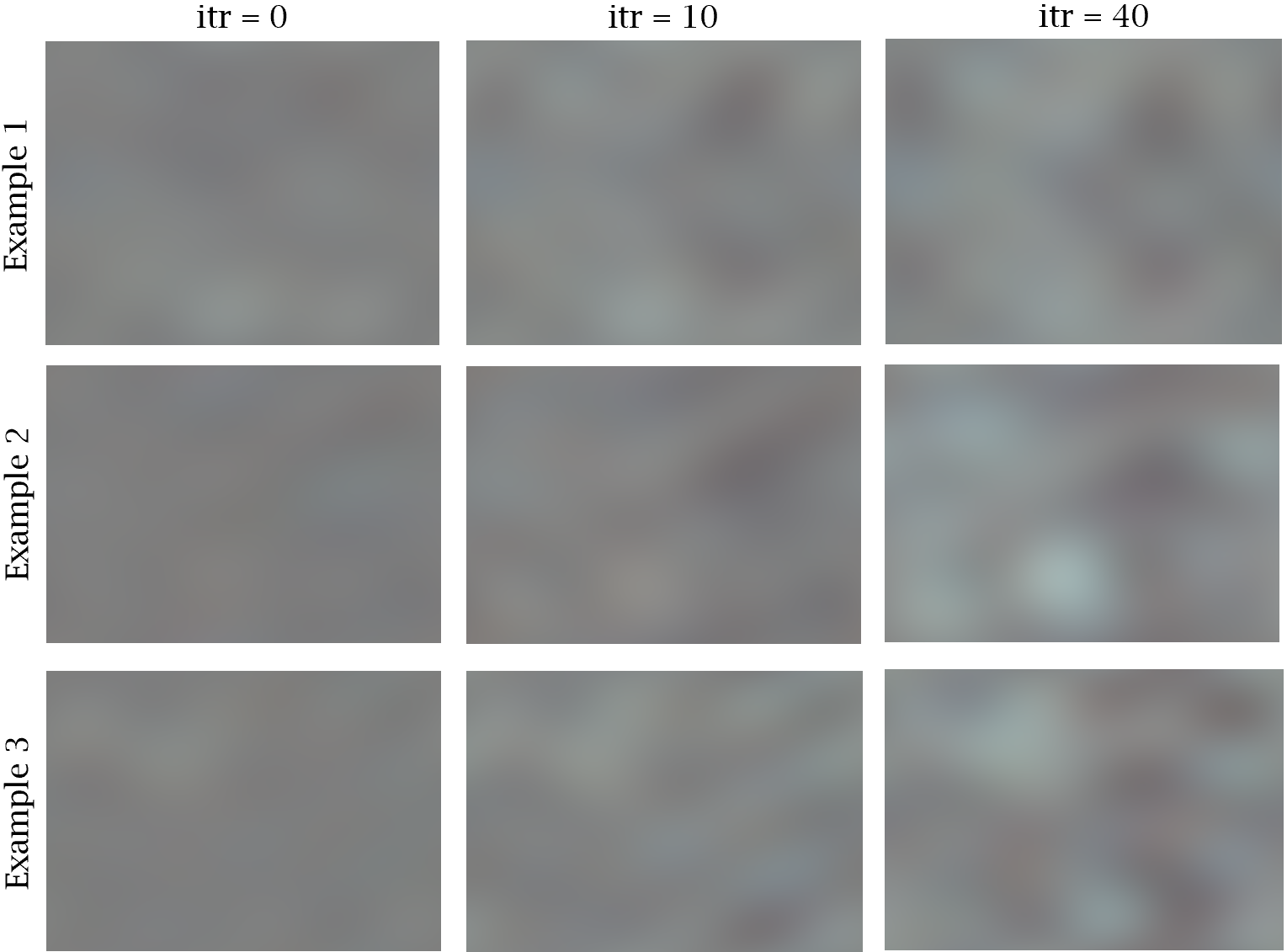}
    \caption{Examples from different image sequence: Caustic pattern being learned progressively as the iteration increases.}
    \label{fig:caustic_evo_viz}
  \end{subfigure}
  \vfill 
  \vspace{0.5cm}
  \begin{subfigure}{0.88\linewidth} 
    \includegraphics[width=\textwidth]{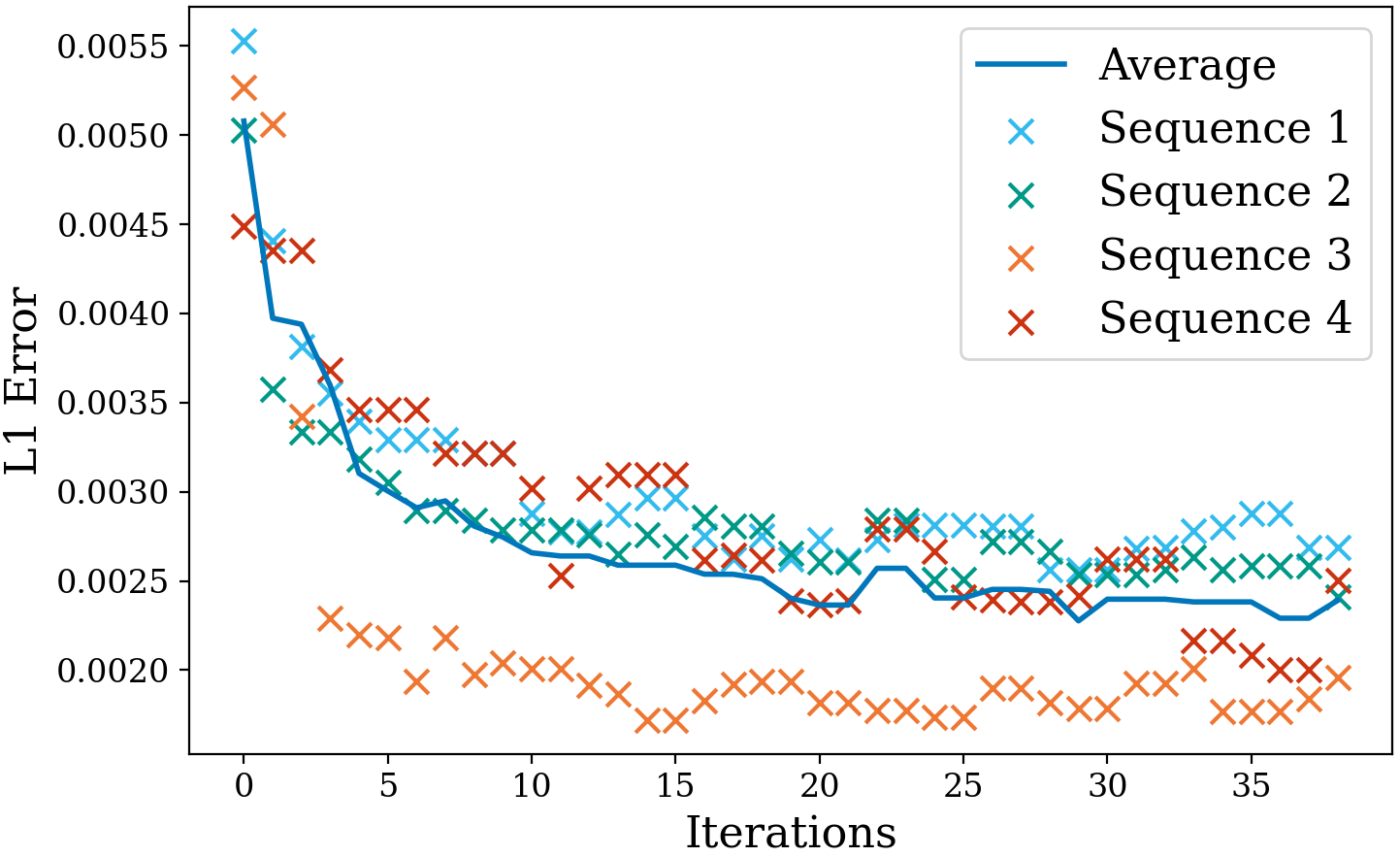}
    \caption{Convergence of our proposed recurrent framework.}
    \label{fig:cau_evo_plot}
  \end{subfigure}
  \caption{Iterations start after warming up the model with a vanilla \ac{3DGS} training pipeline. The caustics are learned progressively.}
  \label{fig:iteration}
\end{figure}

The scaling law has played a critical role on a variety of computer vision tasks in the past decade, that with larger amount of data and model size, the model performance gets significantly improved. Deep learning approaches to caustic removal seek to train on a sufficiently large dataset with labelled data and generalise when provided images, under potentially different conditions and at new locations. However, the community has seen much less progress in scaling up underwater computer vision studies, as the real-world data are extremely expensive to collect and need expert input to label. The \ac{SOTA} data for learning caustic is RCaustic~\cite{seafloorinv2023}, which contains 712 pairs/triplets of images from 7 scenes with annotated caustic patterns and contours. However, for the monocular camera setup, neural networks trained on such a dataset perform poorly when transferring to novel data collected from the field. In Fig.~\ref{exp:deep}, we show that the visual appearance gap between the training data and real-world data, and display an example of the constant failure of pre-trained neural networks when domain gap presents. Our proposed RecGS approach does not suffer the same drawbacks, as it does not rely on training on a labeled dataset to generalize to novel images.

\subsection{Visualization and Comparison with 2D Filtering Method}\label{subsec:filter}
The results on different image sequences are visualized in Fig.~\ref{exp:viz}. Due to the visual nature of this study, viewing the video comparison on our github is highly recommended. The first column shows the original images as input. We highlight the same area with the most significant caustics from one frame to another in red dashed boxes. The second column shows the results rendered from our recurrent \ac{3DGS} framework. By comparing the boxed area in two different frames of the same sequence, we can see that the color and illumination are consistent. The third column shows the residual, which is the difference between the input image and the rendered image. By doing the low-pass filtering on the residual we get the caustic, as shown in the fourth column. In comparison, the motion-compensated filtering method (fifth column) performed well on certain sequences, e.g. Sequence 2, but worse than ours on the other sequences, as inconsistent illuminations can still be observed in the boxed area.

\subsection{How effective are the recurrences?}\label{subset:recurrences}

According to Algorithm~\ref{algo:recgs}, once Eq.~\ref{eq:vanigs} is optimized with vanilla \ac{3DGS}, the 3D Gaussian is considered initialized. Then the algorithm enters the recurrent stage. In each iteration, the caustic is first decomposed from the residual and the 3D Gaussian is optimized for 1000 steps with Eq.~\ref{eq:reoptim}. The way caustics evolve with iterations is shown in Fig.~\ref{fig:caustic_evo_viz}. In the $0^{th}$ iteration, that is, trained with only vanilla \ac{3DGS}, we can only see a faint imagery of caustic. After 10 iterations, we begin to see a clearer pattern of the caustic. After the model has converged in 40 iterations, we see a clear caustic pattern.
\par We plot the $L_1$ error of the caustic returned between two consecutive iterations to support our observation above. A low error means that the model has converged. In Fig.~\ref{fig:cau_evo_plot}, we can see an average trend of error declining, especially in the first 5 iterations, and swiftly converging in less than 40 iterations.

\section{LIMITATIONS}
\label{sec:limit}
\begin{figure}[h]
  \centering 
  \begin{subfigure}{0.48\linewidth} 
    \includegraphics[width=\textwidth]{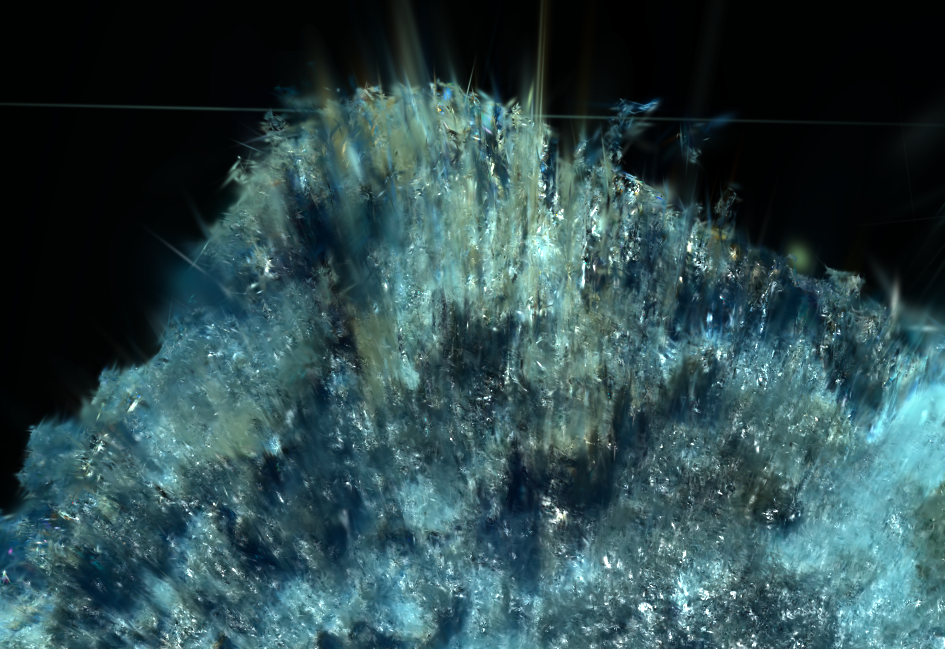}
    \caption{Gaussians build with sparse top-down views fail when viewing from novel side-views.}
    \label{exp:lima}
  \end{subfigure}
  \hspace{0.02\linewidth}
  \begin{subfigure}{0.43\linewidth} 
    \includegraphics[width=\textwidth]{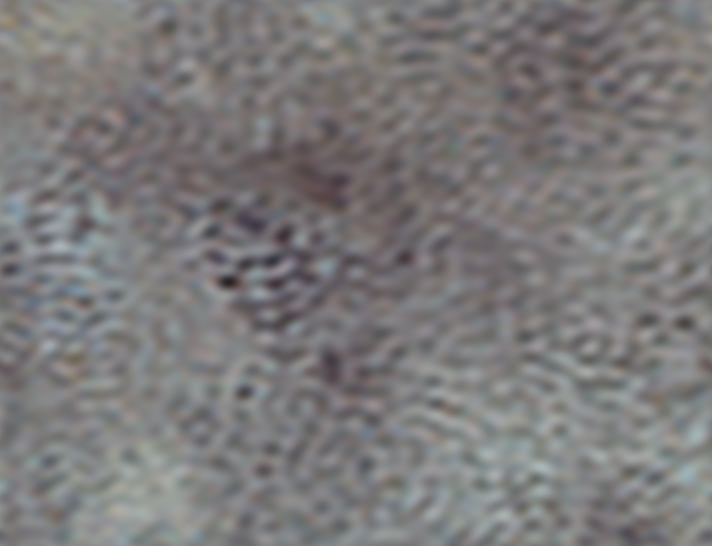}
    \caption{Undesired caustics due to improperly selected hyper-parameter.}
    \label{exp:limb}
  \end{subfigure}
  \caption{Limitations of our method.}
  \label{fig:lim}
\end{figure}

\subsubsection{Sparse-View 3D} The 3D structure built with our \ac{3DGS} has limited capacity to extrapolate beyond viewing angles provided during training. It can fail when evaluating from a novel side-view, when the training data consists only of top-view images, as shown in Fig.~\ref{exp:lima}. In other words, our method so far works well on top-down views similar to the training views collected in our dataset. This limitation is not inherent to our approach, and can be generally observed in NeRF and \ac{3DGS}-based methods. In this paper, we are not able to draw the conclusion that removing caustic helps build a better 3D representation for novel-view photorealistic rendering. The key reason can be that the top-down observations from a robotic setup is not sufficient to constrain 3D geometry. When more advanced methods novel-view synthesis methods that improve generalization beyond the training images emerge, we can integrate them seamlessly into our recurrent framework and also achieve photorealistic underwater reconstruction.

\subsubsection{Parameter Tuning} The frequency threshold in the low pass filtering with \ac{FFT} depends on hand tuning. Choosing an improperly high threshold leads to poor visual results as shown in Fig.~\ref{exp:limb}. Unknown of the proper frequency range can lead to potential failure. 

\section{CONCLUSIONS AND FUTURE WORK}
\label{sec:conclusions}
\par This paper proposes a framework that recurrently removes caustic effects from underwater images. \ac{3DGS} and 2D low-pass filtering are employed in each iteration to build a illumination-consistent 3D representation of the scene and remove caustic from the residual. The experiments are carried out in comparison to different strategies such as joint-optimization, deep learning, and 2D filtering approaches, and the results show that our method provides better visual results. Overall, from this project, we learn that by designing a system that wraps around 3D scene representations, we can learn and recover complex illumination effects with 3D visual consistency. Future work includes building better 3D reconstruction from sparse views that allows not only caustic removal, but also improves novel view rendering. In addition, we hope that our proposed recurrent \ac{3DGS} can inspire and be applied to different applications with challenging imaging effects.

\section*{ACKNOWLEDGEMENTS}
\label{sec:acknowledgements}
This work is supported by NOAA NA22OAR0110624.



\renewcommand{\bibfont}{\normalfont\small}
\printbibliography
\end{document}